# Scaling Semantic Categories: Investigating the Impact on Vision Transformer Labeling Performance

Harrison Muchnic
hem9984@nyu.edu

Anthony Lamelas
anthony.lamelas23@gmail.com

New York University Tandon School of Engineering
6 MetroTech Center, Brooklyn, NY 11201

## Abstract

*This study explores the impact of scaling semantic categories on the image classification performance of vision transformers (ViTs). In this specific case, the CLIP server provided by Jina AI is used for experimentation. The research hypothesizes that as the number of ground truth and artificially introduced semantically equivalent categories increases, the labeling accuracy of ViTs improves until a theoretical maximum or limit is reached. A wide variety of image datasets were chosen to test this hypothesis. These datasets were processed through a custom function in Python designed to evaluate the model's accuracy, with adjustments being made to account for format differences between datasets.*

*By exponentially introducing new redundant categories, the experiment assessed accuracy trends until they plateaued, decreased, or fluctuated inconsistently. The findings show that while semantic scaling initially increases model performance, the benefits diminish or reverse after surpassing a critical threshold, providing insight into the limitations and possible optimization of category labeling strategies for ViTs.*

## 1. Introduction

Computer vision has quickly evolved into a transformative field, allowing machines to understand, interpret, and respond to visual data. Its real-world applications include facial recognition, medical imaging, autonomous vehicles, and more. Computer vision technologies have pushed the limits of artificial intelligence in real-world scenarios.

The goal of computer vision systems has always been to achieve higher accuracy in processing visual information. This began by relying on the manual creation of features and set approaches to analyze visual data and eventually evolved to convolutional neural networks (CNNs) and ViTs. Although CNNs can achieve high levels of accuracy when performing tasks such as object detection, image classification, and segmentation, they face limitations when working on larger scales and when capturing global contexts (Takahashi et al. 2024). This occurs because of their localized convolutional operations. However, ViTs address these shortcomings by treating images as sequences of patches and using self-attention mechanisms to model both local and global dependencies (Takahashi et al. 2024).

Achieving higher accuracy with ViTs often comes at the cost of increased computational requirements. These challenges have led to research on how to improve transformers without compromising their efficiency. This paper focuses on increasing the accuracy of computer vision classification systems by exponentially scaling semantic categories on various datasets.

## 2. Methodology

In order to find out how the addition of semantic categories affects classification accuracy, we measured the classification accuracy with only the base categories. We then added a new set of redundant categories and re-measured the accuracy until it was no longer increasing. The accuracy was measured using the following equation:

$$Accuracy\ Percentage = \frac{Correct\ Classifications}{Total\ Classifications} \times 100 \quad (1)$$

The datasets were gathered using the Internet and resources such as Kaggle. We specifically looked for datasets that provided an answer key or had the correct classification located in the file path. After downloading the image datasets as zip files, the classifications were then measured by running Jina AI's CLIP server. This was performed by using a tool we created, allowing the user to run the server on any Unix command line (Muchnic 2024).

After running the zip file using the labeling tool, a results CSV file is generated with two fields: the file path of the image and the classification. In order to measure the accuracy of these results, custom Python functions were created to parse the file path of each image and compare it to the classification field. The functions accepted a Python Dictionary mapping the correct classification to a list of the semantically equivalent categories. The accuracy was then measured using the equation above (1). This process was repeated until the accuracy stagnated, decreased, or fluctuated.

The same keywords were used to create the semantically equivalent categories to keep the experiment consistent. These were:

(Let C be the correct classification) C, Cs, Large C, Large Cs, Small C, Small Cs, Medium C, Medium Cs, Broad Category C, Broad Category Cs.

If the accuracy was still increasing, more equivalent categories were created by the researcher. After each iteration of the research process, the accuracy and accuracy change were measured to prepare for analysis.

Our methodology was well-suited for this research because it allowed for a systematic and direct analysis of how each addition of semantic categories influenced classification accuracy. By using consistent keywords to generate the new categories, uniformity, and minimal variability were ensured.

## 3. Results

In order to measure each change in the accuracy of the image classification, we measured the accuracy every time a new semantic category was shown. The trends of each measurement can be shown below, with an overall increase shown in every dataset (Figure 1). For the large majority of tested datasets, accuracy increased after the first addition of a semantic category. After the second addition of a semantic category, the accuracy increased for all of the datasets.

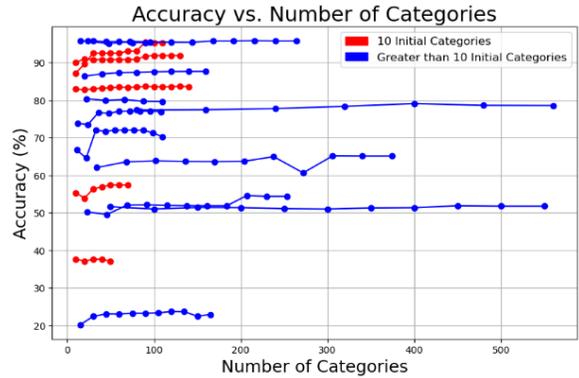

Figure 1: Accuracy Trends Across all Datasets

To understand the trends, the accuracy increase (from the original measurement with base categories) for each iteration where a new semantic category was added was averaged out and plotted (Figure 2). The graph shows that in almost all cases, the accuracy was greater than the original, especially indicating that the initial additions of semantic categories greatly affected accuracy.

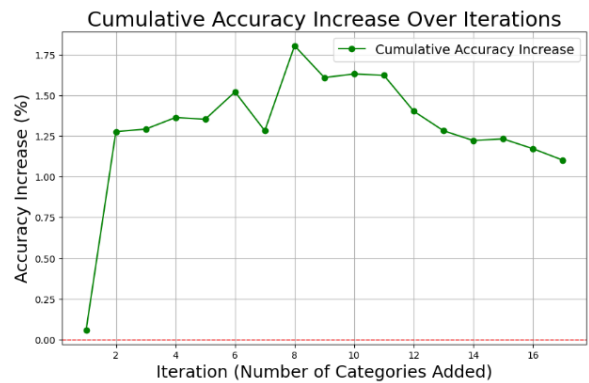

Figure 2: Accuracy Increase per Iteration

To focus on each specific addition of a new category, the average accuracy change from one category addition to the next was examined. The data (Figure 3) shows that initially, there is a strong increase, but at a certain point (around the tenth new category), the change plateaus.

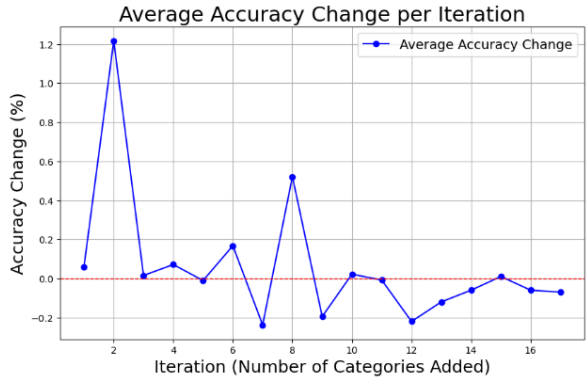

Figure 3: Accuracy Change per Iteration

When examining sets with small amounts of initial categories (defined as less than 10), the results were not nearly as clear. As shown in the bar graph below (Figure 4), there is on average only around a one percent overall increase, which is around half the increase of the sets with more than ten initial categories. It should also be noted that there was no increase in half of the small datasets we analyzed.

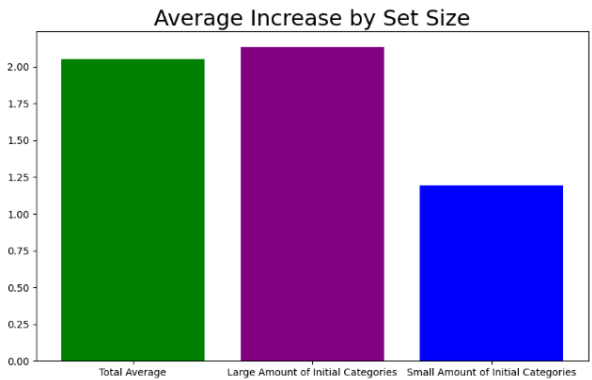

Figure 4: Average Cumulative Accuracy Increase by Amount of Initial Categories

The figure below (Figure 5) shows the peak accuracy change for each of the analyzed datasets. The number next to the dataset name depicts the number of initial starting categories. As you can see, the Flowers10 dataset showed the largest increase, with almost all of the datasets with initial category amounts greater than ten having increases. However, of the four datasets with initial categories less than ten, two of them decreased in accuracy, those being Card4 and Veg7.

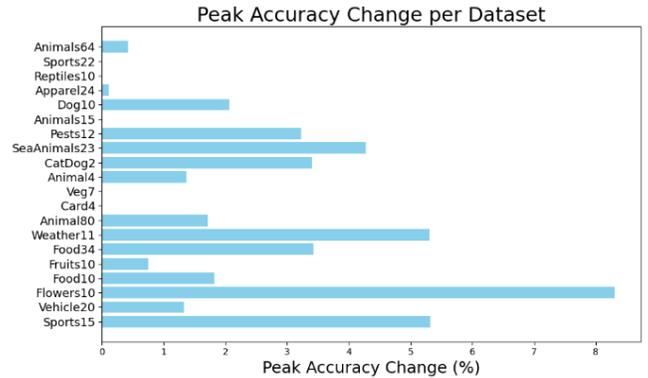

Figure 5: Peak Accuracy Increase per Dataset

Our analysis of the classification accuracy changes when introducing semantically equivalent categories showed a direct correlation between the addition of categories and accuracy increase. This can be seen when further examining the accuracy increase per iteration, specifically when comparing to the accuracy when using the original categories (Figure 6).

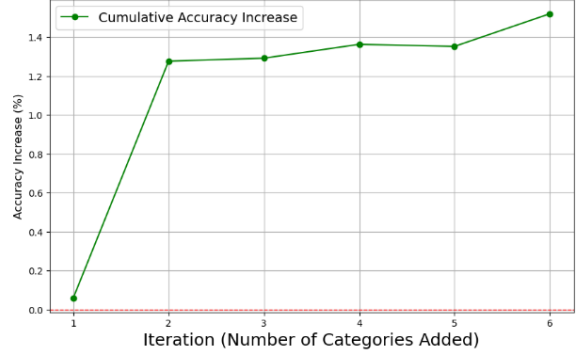

Figure 6: Accuracy Increase per Iteration (Through 6 Iterations)

However, after the initial increase in accuracy, a plateau is eventually reached where either the increase ends or the accuracy begins to decrease. This directly proves our initial claims and can be visualized when further examining the accuracy increase per iteration, specifically when comparing to the accuracy of the original categories after the sixth iteration (Figure 7).

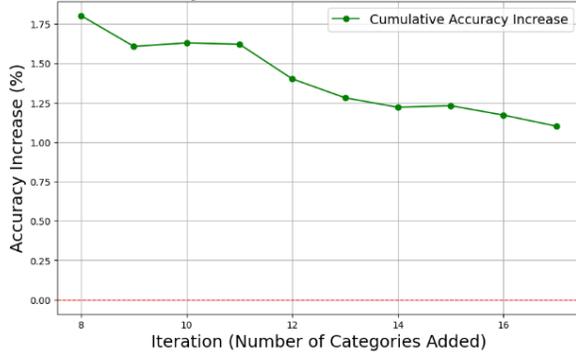
Figure 7: Accuracy Increase per Iteration (After 6 Iterations)

### 3.1. Limitations and Recommendations for Future Research

One possible limitation may be the fact that we only examined a limited number of datasets due to the computational expense of ViTs. This may not fully represent all possible outcomes. Another possible limitation could be that the semantic categories were human-generated, which may introduce potential biases. The results are also specific to the JINA model that we used and may not be fully consistent for all ViTs.

Future research could involve scaling semantic categories in architectures other than ViTs, such as hybrid models (combining convolutional layers and transformers) or CNNs. A future experiment could also extend what we performed onto more diverse datasets containing more images, greater image diversity, etc. Other next steps could include researching the optimal number of redundant categories automatically and/or investigating the mathematical theory behind the plateauing factors.

### 4. Conclusion

This research represents a significant step in understanding the impact of scaling semantic categories in ViT image classification. By systematically introducing semantically equivalent categories, we observed a clear trend of accuracy improvement until a certain plateau or decline was reached. This finding highlights the potential for semantic scaling to optimize classification performance in complex datasets while acknowledging its diminishing returns when pushed beyond a critical threshold.

However, this study has its limitations. The dataset selection and reliance on manual category creation may have influenced the results. Additionally, the computational expense restricted the number of datasets that could be analyzed.

However, our work provides a foundation for future research in this area. Such research could focus on testing a broader range of datasets or using classification tools other than ViTs. Practical applications, such as medical imaging and autonomous vehicles, could also benefit from strategic semantic scaling to improve accuracy in real-world environments. This study looks to further category labeling strategies, bridging the gap between theoretical models and real-world implementations.